# Creating a Large Multi-Layered Representational Repository of Linguistic Code Switched Arabic Data

**Mona Diab, Mahmoud Ghoneim, Abdelati Hawwari,
Fahad AlGhamdi, Nada AlMarwani, Mohamed Al-Badrashiny**

Department of Computer Science
The George Washington University
{mtdiab, mghoneim, abhawwari, fghamdi, nadaoh, badrashiny}@gwu.edu

**Abstract**

We present our effort to create a large Multi-Layered representational repository of Linguistic Code-Switched Arabic data. The process involves developing clear annotation standards and Guidelines, streamlining the annotation process, and implementing quality control measures. We used two main protocols for annotation: in-lab gold annotations and crowd sourcing annotations. We developed a web-based annotation tool to facilitate the management of the annotation process. The current version of the repository contains a total of 886,252 tokens that are tagged into one of sixteen code-switching tags. The data exhibits code switching between Modern Standard Arabic and Egyptian Dialectal Arabic representing three data genres: Tweets, commentaries, and discussion fora. The overall Inter-Annotator Agreement is 93.1%.

**Keywords:** Code Switching, Computational Linguistic Resources, Annotation, Sociolinguistics.

## 1. Introduction

Linguistic Code Switching (CS) is a common practice among multilingual speakers in which they switch between their common languages in written and spoken communication. A Spanish-English blog entry illustrates this: *"She told me that mi esposo looks like un buen hombre."* (*"She told me that my husband looks like a good man"*). CS is typically present on the inter-sentential, intra-sentential (mixing of words from multiple languages in the same utterance) and even morphological (mixing of morphemes) levels. This phenomenon can be observed in different linguistic levels of representation for different language pairs: phonological, morphological, lexical, syntactic, semantic, and discourse/pragmatic switching.

For multilingual speakers, CS is pervasive in their spoken and in informal written genres such as email and web blogs. CS presents serious challenges for language technologies, including parsing, machine translation (MT), automatic speech recognition (ASR), semantic processing, and information retrieval (IR) and extraction (IE). Techniques trained for one language quickly break down when there is input from another. Being able to predict/identify probable switch points, as well as which dialect/language a speaker is switching to, enables applications to adapt their models.

A major barrier to research on CS has been the lack of large, consistently and accurately annotated corpora of CS data. In the shared task for "Language Identification in Code-Switched Data" (Solorio et al., 2014), the first set of annotated data was created which focused on social media and covered four language pairs: Modern Standard Arabic - Dialectal Arabic (MSA-DA), Mandarin - English (MAN-EN), Nepali - English (NEP-EN), and Spanish - English (SPA-EN).

In this work we present our effort to build a large repository of CS data that will cover multiple language pairs and dialects. We started by focusing on Arabic Language. Arabic is a Semitic language spoken by over 300M people worldwide. CS between MSA and DA is widespread among native speakers of Arabic. MSA is the language of education used in formal speeches and settings, while DA is the everyday spoken variant; even minimally educated Arabic speakers speak two languages. While there is considerable lexical overlap between MSA and DA, a significant number of MSA items have taken on senses that are quite different in DA. Such divergence causes serious problems for automatic analysis. For example the phrase "كتب كتابه/katab kitAbuh/wrote book-his"[1] literally means "wrote his book" in an MSA context while the more dominant meaning in an Egyptian dialect context is "he got married". Arabic-English (Ar-En) CS occurs mostly between DA and English, which go beyond technical term borrowings and *nonces*,[2] where we can see an English phrase modified by Arabic morphology and/or phonology. The following example illustrates morphological DA-English CS:

**EGY-DA**: هتفرمت الديسك ولا هتستنا شوية

**Transcription**: *ha-tofaromat Al-disok wal~A ha-tesotan~aA $uway~ap*

**Gloss**: *will-you-format the-disk or will-you-wait a-bit*

**English Translation**: "Will you format the disk, or would you rather wait a bit?"

The first VP "*ha-tofaromat Al-disok*" is an English phrase with Arabic morphology and phonology illustrating a typical inter-linguistic CS phenomenon.

This paper is organized as follows: Section2 reviews related work. Annotation standards and guidelines are discussed in sections 3 and 4 respectively. The annotation process is detailed in section 5. The current status of our repository is presented in section 6 followed by a discussion in section 7. Finally conclusions and future work are discussed in section 8.

---

[1] Examples are in the form: "Arabic Script / Buckwalter Transliteration / English meaning"
[2] Words coined 'for the nonce' which may later enter the language.

## 2. Related Work

Theoretical linguistic research on CS has claimed it to be a structurally coherent, rule-governed linguistic behavior. While many proposals have been made attempting to define this rule system (MacSwan, 1999), these have yet to be empirically verified on multiple language pairs. For example, it is not clear whether CS involves the integration of two separate grammars (Cook, 1992; Grosjean, 1989) or a single grammar that unified the two (Lederberg & Morales, 85; Myers-Scotton, 1993; Muysken, 2000). Recent work posits two hypotheses (MacSwan 1999, 2000, 2005; Chan, 2003, 2008; Gelderen and MacSwan 2008): (1) nothing constrains CS but the requirements of the two grammars involved; and (2) CS is constrained by the same rules that govern monolingual speech. Others believe that CS operates within a system that specifies the syntactic environments in which language alternation may or may not occur. For example, Myers-Scotton's (1993) Matrix Language Frame (MLF) model proposes that the Matrix Language supplies the morpho-syntactic framework and the Embedded Language may optionally insert particular switched, primarily content, elements into that framework.

Albirini et. al. (2011) have shown that Arabic-English CS exhibits switches between smaller constituents such as Noun Phrases rather than larger ones such as subordinate clauses. Others (Bassiouney, 10; Dashti, 07; Redouane, 2005) have claimed that CS point occurrences are bound both morphologically and syntactically. We see several of these studies in the sociolinguistics and theoretical literature; however, no serious computational linguistics application has exploited such knowledge due to the lack of a suitably annotated training corpus.

With few exceptions, language technology research has not addressed issues of CS. The exceptions, however, do show that CS must be addressed in order to obtain performance similar to monolingual speech processing. Lyu et al. (2006) found that building a unified acoustic model of the regional dialects to be detected, a bilingual pronunciation model, and a Chinese character-based tree-structured search strategy improved ASR performance significantly. Solorio & Liu (2008) found that CS poses a serious challenge to part-of-speech tagging: while monolingual taggers reach >96% accuracy, English taggers tested on Spanish-English CS data obtain only 65% accuracy. Chiang et al (2006) similarly reported that POS taggers trained on MSA dropped from 96.15% to 77% accuracy when run on data including CS to Arabic dialects. The lack of large labeled CS corpora seriously hinders the development of language tools that approach monolingual tools' levels of performance. Annotated corpora for multiple language pairs are needed to provide training data needed to build these tools. Some initiatives to create CS annotated corpora have been reported (Li et al., 2012; Dey and Fung, 2014; Maharjan et al., 2015) and the first shared task on language identification in CS data took place recently (Solorio et al., 2014)

## 3. Transcription and Annotation Standard

A common transcription and annotation standard is crucial to sharing the data collected and annotated. This standard should allow interoperability for cross-language pair comparisons. We have developed an XML encoding schema that supports four annotation levels: Document Level, Word-level, CS points and Syntactic level. For the resource presented in this paper, we only fully fulfilled the first three levels and partially the fourth one.

Document annotation includes all meta-data information available describing the source of the document, the languages involved, any speaker information available (age, gender, education, language background, regional origin), and genre. For every word, the language is identified. In the case of mixed language words, the language for each morpheme is identified separately. The part-of-speech (POS) of the word is also assigned. The CS points are identified by the change of the word language tag.

| #  | POS-Tag   | POS categories                                                                                                       |
|----|-----------|----------------------------------------------------------------------------------------------------------------------|
| 1  | NOUN      | Noun, Number NOUN, Quantitative Noun                                                                                 |
| 2  | VERB      | Verb, Pseudo Verb                                                                                                    |
| 3  | ADJ       | Adjective, Comparative Adjective, Number Adjective                                                                   |
| 4  | PRON      | Pronoun                                                                                                              |
| 5  | NOUN_PROP | Proper Noun                                                                                                          |
| 6  | PART      | Particles (Vocative Particle, Restriction Particle, Future Particle, Negation Part, Focus Part, Interrogative Part) Sub Conjunction |
| 7  | PREP      | Preposition                                                                                                          |
| 8  | ADV       | Adverbs, Relative Adverbs, Interrogative Adverbs                                                                     |
| 9  | DET       | Demonstrative, Demonstrative Pronoun                                                                                 |
| 10 | CONJ      | Conjunction                                                                                                          |
| 11 | INTERJ    | Interjection, Exclamation Pronoun                                                                                    |
| 12 | ABBREV    | Abbreviation                                                                                                         |
| 13 | MWE-Com   | A part of a multiword expression                                                                                     |
| 14 | NE-Com    | A part of a named entity construction                                                                                |

Table 1: POS tag set

## 4. Annotation Guidelines

We have developed two versions of Code Switching Annotation Guidelines; the first version is a generic one with common coarse-grained tag set to allow for cross-linguistic analysis, and the second uses fine-grained tag set tailored for the Arabic language and its dialects. For each annotation guideline, we created a detailed version for Gold (in-lab) annotation and a simplified version suitable for crowd-sourcing annotation.

The tagging level of our annotation is the token, taking into account the context of the entire sentence. We assumed that an efficient corpus for CS should have information about the POS for each word, and mark the orthographic error, if any. Therefore, our annotation guidelines include three different tag lists for word/token level annotation: **a) CS tags**, **b) Orthographic Error tags**, and **c) POS tags**. Tables 1 and 2 show POS and CS tag sets respectively. The errors tag list includes two tags: a) "**Typo**": to indicate different types of typos like misspelling, splits and merges, and b) "**Correct**" which is assumed to be the default case.

| | # | Label | Description |
|---|---|---|---|
| linguistic | 1 | MSA | Modern Standard Arabic: words that can only be used as MSA (e.g. "منذ/muno*u/since") or shared words that are MSA in the given context. |
| | 2 | DA | Dialect Arabic: words that can only be used as Dialectal word (e.g. "معلهش/maEalih$~/take-it-easy") or shared words that are DA in the given context. |
| | 3 | Ambiguous | Ambiguous words: semantic cognates, shared words that can equally be considered MSA or DA in the given context. Usually occurs when the sentence is too short to detect the context or when the word is located between two different contexts and it is not clear to which one it belongs. |
| | 4 | MA | Mixed Arabic: cases where an MSA inflectional morpheme (affix or clitic) is attached to a totally dialectal word (e.g. "سيزعل/sa_yazoEal/will-be_displeased", the MSA future marker "sa" is used instead of the DA future marker "ha") |
| | 5 | FW | Foreign Words: words that is not originally part of the Arabic language (e.g. "أبسليوتلي/>abosoluwtoliy/absolutely" |
| | 6 | MF | Mixed Foreign: cases where an Arabic inflectional morpheme (affix or clitic) is added to a foreign word (e.g. "هتفرمت/ha-tofaromat/will-you-format") or a foreign inflectional morpheme is added to an MSA or DA word. |
| | 7 | NE | Named Entity: a name of a unique entity such as names of persons, geographical locations, organizations, events, etc. |
| Non linguistic | 8 | UNK | Unknown |
| | 9 | Latin | Words written in English letters |
| | 10 | URL | Web links and emails |
| | 11 | Punctuation | All punctuation characters |
| | 12 | Number | Numbers and digits |
| | 13 | Diacritics | Diacritics |
| | 14 | Emotion | Symbols that represent emotions |
| | 15 | Sound | String of letters that represent sounds (e.g. "hahaha") |
| | 16 | Other | All tokens that cannot be classified to the other 15 labels |

Table 2: Summary of the 16 CS-Labels for data categorization.

The guidelines have been refined over several iterations, making use of the annotation disagreement analysis over thousands of tokens annotated iteratively by four native speaking annotators. Although it is built upon a solid linguistic background, our guidelines are meant to be as easy as possible by reducing the technical terms and adding tens of illustrative examples and detailed explanations for real annotated data. In addition to the guidelines document, we offer a fully annotated sample repository.[3]

## 5. Annotation Process

### 5.1 Annotation Team

We have a native speaking team of three annotators and one lead annotator. Most of the annotators have a linguistic degree. A three weeks training period with annotation guidelines is mandatory for each annotator. Face-to-face team meetings are held on a weekly basis to discuss annotation findings and feedback.

### 5.2. Data Harvesting

The data harvested so far comes from three resources: LDC Egyptian Arabic Treebanks parts 1-8 (ARZ) (Maamouri et al., 2012), the Arabic online commentary dataset (AOC) (Zaidan and CallisonBurch 2011) and

---
[3] The annotation guidelines and example repository are available at http://care4lang1.seas.gwu.edu/cs

4230

Twitter (TWT). ARZ data comes mainly from discussion forums. AOC is reader commentaries that were crawled from an Egyptian Newspaper called "Al-Youm Al-Sabe". TWT data is crawled from some Egyptian public figures' Twitter accounts.

## 5.3. Data Preprocessing

A preprocessing pipeline is developed to prepare data for annotation. First, raw text data is extracted from sources and different cleaning steps (such as handling non-standard characters) are carried out using the Smart Preprocessing (Quasi) Language Independent tool (SPLIT) (Al-Badrashiny et al. 2016). Then Automatic Identification of Dialectal Arabic (AIDA2) tool (Al-Badrashiny et al. 2015) is used to assign initial automatic tagging for highly confident data categories (label types 9 through 15 in table 2) in addition to named entities (label type 7). Finally, the preprocessing pipeline puts the data in the format acceptable by the annotation application.

## 5.4. Gold In-lab Annotation

Initially we started using Google sheets for bootstrapping the in-lab annotation process. This has the advantage of accommodating the dynamic nature in terms of requirement changes and design for low overhead cost. As we go along the annotation process, the need for a specialized annotation tool that can streamline the management of large-scale annotation became apparent. We developed a web-based CS annotation tool that facilitates managing multiple CS annotation tasks. The tool offers several levels of management and produces quality control measures and annotation statistics. The tool is a typical three-tier web application. The data tier stores meta-data in PostgreSql database in addition to the raw and annotated data files, which are stored on a file server. The Logic tier consists of PHP scripts interact with Apache web server. It implements all functionalities provided by the system to the different types of users. The web server sends requests to the database server through a secured tunnel. The presentation tier is browser independent, which enables accessing the system from many different clients. It also supports multiple encodings to enable multilingual annotation. It provides intuitive Graphical User Interface tailored to each user type. This architecture enables multiple annotators to work on different tasks simultaneously. On the other hand, the administrator manages only one central database. The tool integrates with different pre-processing tools (such as SPLIT and AIDA2) and supports exporting the annotation in the standard format. Figure 1 shows system architecture. The system has built-in functionality to manage annotation assignment overlap necessary for calculating Inter-Annotator Agreement (IAA) per task. It also provides useful progress reports and statistics.

### 5.4.1 Types of Users

Three types of users have been considered in the design of the tool: **Super-user, Lead Annotator,** and **Annotator**. Each type of users is provided with different kinds of privileges and functionalities in order to fulfill their tasks.

**Super-user**: There is only one super-user account for all dialects/languages. The super-user manages users' accounts, data import and export in addition to monitoring the overall performance of the system.

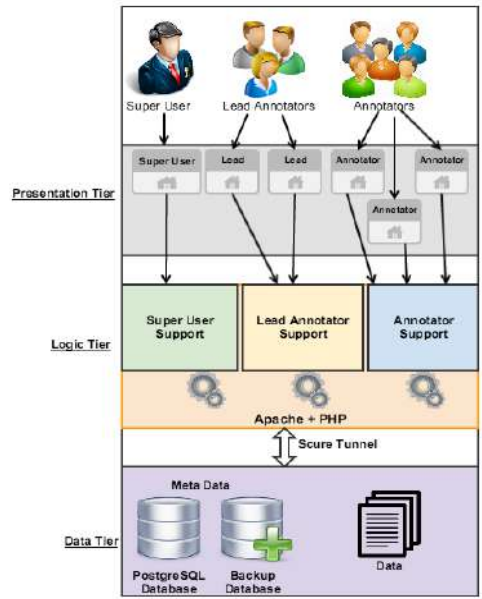

Figure 1: Annotation Tool System Architecture

**Lead Annotator**: There is one lead annotator account for each dialect/language. The lead annotator manages annotation task assignment, monitors status and progress, reviews and grades annotators' work and produces different quality measures. The system enables the lead annotator to reject submitted work that does not meet the assessment criteria as well as provide comments and feedback to the annotator to re-annotate the rejected work. The lead annotator can specify the percentage of assigned data overlap between annotators, which is used to calculate IAA.

**Annotators**: The system enables annotators to access their assigned tasks, annotate words in context, save partially finished tasks, check lead annotator feedback and grading on submitted tasks, re-annotate rejected tasks and access online help and guidelines. The interface uses color-coding to reflect useful information and status. For example, 'named entity' tagged words are highlighted in purple, while words with label types 9 to 15 in table 2 are highlighted in orange. Words that are annotated are displayed in blue while words that are not yet annotated are black.

### 5.4.2 Database Design

The system uses a relational database to store and manage all meta-data. These data falls under one of the following categories:

**Profiling information:** It contains information about the different registered users of the system including their assigned role (i.e. annotator, lead annotator or super user), their login information as well as the dialect and languages for each one of them. It also contains information about different registered language/dialect pairs.

**Annotation Information:** This is the core part of the database. It contains all meta-data related to the annotation work such as annotations completed by each annotator and temporarily saved annotations.

**Assessment Information:** It contains information about 1) Task-Annotator assignment: including which tasks are assigned to each annotator and how many tasks have been annotated and submitted, the total number of assigned units (post, tweets), percentage of data annotation overlap to facilitate inter-annotator agreement calculations, number of annotated units, genre type, etc.; 2) Annotator-Units assignment: including information about each unit (post, tweet) assigned to the annotators such as post-id, genre-id, assignment-id, path of the assigned file;   Finally 3) Language-Unit assignment: which includes information about which unit belongs to which dialect/language.

### 5.5. Crowed Sourcing Annotation

In our effort to leverage crowd sourcing software platforms for soliciting the bulk of our future CS annotation, we used CrowdFlower platform to conduct a pilot experiment on 300 Levantine tweets (2782 tokens, 1898 types). We used the simplified version of the guidelines, which provide a basic description of each tag along with examples on how to perform the task and notes on how to handle typo. To simplify the task, we dropped annotation verification for the automatically identified CS tags (types 9 to 15 in table 2) and focused only on 9 categories (CS tags 1 to 8, and 16 in table 2). The annotators were asked to select the correct label from a drop down menu for the highlighted word in context. Figure 2 shows an example task. The tasks were restricted to Arabic speaking workers. Before conducting the task, workers must obtain at least 75% accuracy in a qualifying quiz composed of a gold annotated set where they have to correctly annotate 15 out of 20 words. During the task, hidden gold data continuously appear in their job, so that we maintain the 75% minimum accuracy. These gold data (total 300) have been annotated manually in-lab by two annotators.

The task ran for 3 days and a total of 54 workers took the quiz but only 8 qualified. Only 5 workers maintained the minimum accuracy requirement. The overall and per tag IAAs, calculated using Fleiss' Kappa (Fleiss, 1971), were very low which indicates that our crowdsourcing setup needs revision and refinement.

### 5.6. Quality Control

To control the quality of the annotation process, at least 10% of the weekly assigned data is anonymously shared between the annotators. On a weekly basis, this overlapped data is used to calculate the IAA for all tags in addition to each single tag. The IAA results is discussed during Annotators' weekly meeting and the annotations of data batches with less than 90% IAA are repeated. If the IAA for a certain class is below 80%, the annotation guidelines for that class is revised for clarity and guidelines are updated accordingly. Part of our quality control plan involves an external advisory board to help with advice and input on strategy and direction. Another additional external mechanism for quality control is the release of our annotations to the community at large to test its usefulness for NLP system development.

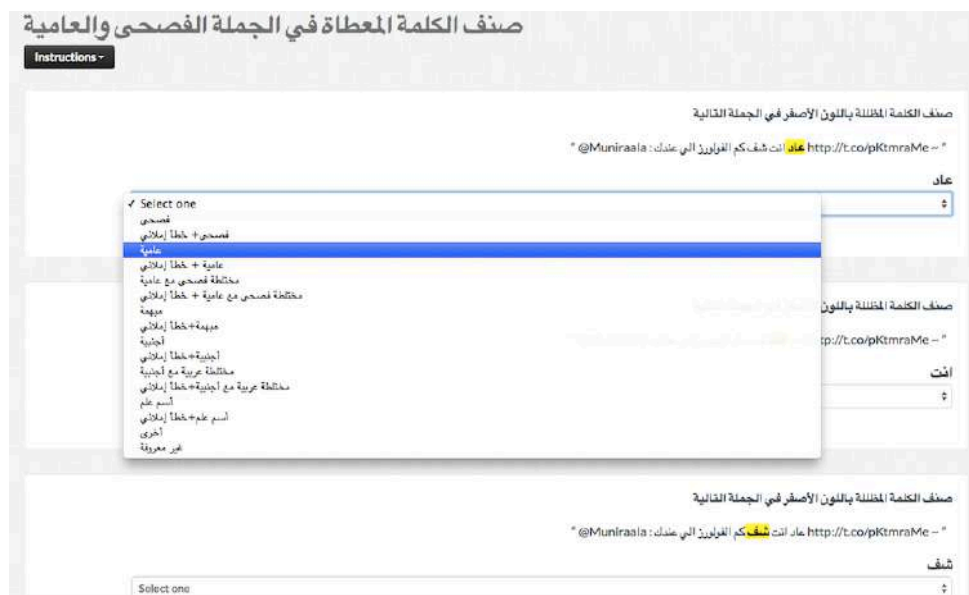

Figure 2: CS Crowd Sourcing Annotation Task using CrowdFlower Platform

## 6. Repository Statistics

The annotation process in the CS project is still in progress. A portion of the annotated data was released to participants of the Code Switching shared task at EMNLP 2014. Currently, a total of 886,252 tokens have been annotated. The average speed of annotation is 792 tokens/hour and the overall Inter-Annotator Agreement is 93.1%. The per-tag IAAs are shown in table 3. The genres of the annotated data are: Discussion Forums, News Commentaries and Tweets. More details about number of annotated tokens, types, and tag distributions are given in table 4.

## 7. Discussion

The analysis of annotators' disagreement is very crucial for the development of the guidelines and the evaluation of repository's quality. While the overall IAA is 93.1%, some per-tag IAA's are low. The "UNK" tag is by definition very annotator dependent. Annotators are required to consult different dictionaries to look up words before deciding that these words are unknown. Accordingly, most of the "UNK" tags are due to typos. The low agreement comes from the differences in annotators' ability to predict the intended word. For "Ambiguous", "MA", "MF" and "Sound" tags, the frequencies of these tags in the corpus are very low; hence, any small disagreement has a large effect on their IAA scores.

The accuracy of assigning "FW" tag depends on the etymological knowledge of the annotator to distinguish between borrowed foreign words that are Arabized long ages ago and became part of the language versus those newly borrowed words that are still considered foreign. Words like "افندم/Afanodim", "باشا/bA$A/Pasha" and "بيه/byh" are considered part of the Egyptian Dialect although they are borrowed from Turkish language.

While the use of AIDA2 to assign initial automatic tags boosts the annotation speed, we found some errors due to annotators' tendency to keep initial automatic tags. For example, the phrase "25/يناير ٢٥/yanAyr/25 January" would have the initial automatic tagging "number NE", while in some contexts it might refers to the Egyptian revolution and should be tagged as "NE NE". Another example is names written in Roman script. These are automatically tagged as "Latin" while it should be tagged as "NE". Other disagreements related to the "NE" tag come from the interpretation of adjectival phrases. For example, the collocation "والله العظيم/w_Allh AlEazym/ I-swear-of_the-God the-sublime" might be tagged as "NE MSA" if the annotator recognizes "العظيم / AlEazym / the-sublime" as an adjective or tagged as "NE NE" if he recognizes it as part of the collocation.

| Label | IAA |
|---|---|
| MSA | 94.83% |
| DA | 92.15% |
| Ambiguous | 28.44% |
| MA | 39.14% |
| FW | 72.61% |
| MF | 75.47% |
| NE | 88.17% |
| UNK | 22.57% |
| Latin | 88.35% |
| URL | 100.00% |
| Punctuation | 99.93% |
| Number | 98.04% |
| Diacritics | 100% |
| Emotion | 100% |
| Sound | 95.61% |
| Other | 98.09% |
| **Overall** | **93.10%** |

Table 3: Per-tag and overall IAAs

| Corpus | Genres | Dialect | Tokens | Types | Tag Distributions |
|---|---|---|---|---|---|
| AOC | News / Commentaries | EGY | 358988 | 67570 | **MSA**:179115, **DA**:121398, **Ambiguous**:148, **MA**:55, **FW**:969, **MF**:2123, **NE**:33158, **UNK**:566, **Latin**:624, **URL**:53, **Punctuation**:17953, **Number**:2445, **Diacritics**:101, **Emoticon**:33, **Sound**:266, **Other**:59 |
| TWT | Tweets | EGY | 206554 | 42884 | **MSA**:132947, **DA**:30476, **Ambiguous**:1077, **MA**:19, **FW**:532, **MF**:1086, **NE**:24386, **UNK**:15, **Latin**:0, **URL**:0, **Punctuation**:0, **Number**:0, **Diacritics**:0, **Emoticon**:0, **Sound**:0, **Other**:15626 |
| ARZ | Discussion Forums | EGY | 84138 | 22228 | **MSA**:17579, **DA**:53084, **Ambiguous**:0, **MA**:3, **FW**:8, **MF**:616, **NE**:5406, **UNK**:0, **Latin**:0, **URL**:31, **Punctuation**:6955, **Number**:414, **Diacritics**:0, **Emoticon**:2, **Sound**:6, **Other**:0 |

Table 4: Statistics of current version of CS-annotated repository

The most frequent tags are the "MSA" and "DA" tags and they are the most interchangeable tags. Most of the disagreement comes from different phonological performance by the annotators. For example, the sentence "أي خبر فيه مصلحة الجزائر يفرح أي مصري حقيقي/Ay xbr fyh mSlHp AljzA}r yfrH Ay mSry Hqyqy/any news in-it benefit to-Algeria would-please any Egyptian real" would be considered DA if the word "يفرح" is read as "يفَرَّح / yfar~aH" or MSA if it is read as "يُفْرِح / yuforiH.". Another source of disagreement is the span of the code switching. In the sample disagreement shown in Table 5, the first annotator assumes the totally dialectal word "مش/m$/Not" is a token replacement of the MSA word "ليست/lyst/Not" and hence the span of the code-switching is only one token, while the second annotator considers the dialectal word as an indicator of a dialectal reading and annotated the narrowest meaningful phrase as dialectal.

| Word | Annotator1 | Annotator2 |
|---|---|---|
| ولكن | MSA | DA |
| أجهزتنا | MSA | DA |
| الجنائية | MSA | DA |
| لأنها | MSA | DA |
| مش | DA | DA |
| خيال | MSA | DA |
| علمى | MSA | DA |
| لم | MSA | MSA |
| تجد | MSA | MSA |
| ولو | MSA | MSA |
| معلومة | MSA | MSA |
| واحدة | MSA | MSA |

Table 5: Sample CS span disagreement.

## 8. Conclusion and Future Work

We presented our effort to create a large Multi-Layered representational repository of Linguistic CS Arabic data. We developed guidelines for annotating and tagging each word in our multi-genre corpus, with 16 code-switching tags, and POS tags. Two annotation protocols have been used within annotation processing; in-lab and crowd sourcing. To validate the annotated data, we applied several quality control measures. The result is a wide-coverage, accurately annotated data that classifies each single word in our repository into one of sixteen code-switching tags. While the main bulk of the annotation so far was carried out using Google Sheets, the annotation tool we developed proved very successful and essential in the management of the annotation process, it is worth noting that the average annotation speeds using the two systems are comparable. So far, we used Egyptian dialectal data. We are currently working on other Arabic dialects; Levantine, Iraqi, Gulf, Moroccan and Tunisian.

## 9. Acknowledgements

We would like to thank the anonymous reviewers for their valuable comments and suggestions. We also thank the in-lab annotators and the CrowdFlower contributors. This work was partly funded by NSF under awards 1205475 and 1205556. The statements made herein are solely the responsibility of the authors.

## 10. References

Al-Badrashiny, M., Elfardy, H., & Diab, M. (2015). AIDA2: A Hybrid Approach for Token and Sentence Level Dialect Identification in Arabic. CoNLL 2015, 42.

Al-Badrashiny, M., Pasha, A., Diab, M., Habash, N., Rambow, O., Salloum, W., Eskander, R. (2016). SPLIT: Smart Preprocessing (Quasi) Language Independent Tool. In the proceedings of the 10th International Conference on Language Resources and Evaluation (LREC'16).

Albirini, A., Benmamoun, E., & Saadah, E. (2011).Grammatical features of Egyptian and Palestinian Arabic heritage speakers' oral production. Studies in Second Language Acquisition, 33(02), 273-303.

Bassiouney, R. (Ed.). (2010). Arabic and the media. Brill.

Benajiba, Y., & Diab, M. (2010). A web application for dialectal arabic text annotation. In Proceedings of the LREC workshop for language resources (LRS) and human language technologies (HLT) for Semitic languages: Status, updates, and prospects.

Chan, B. H. S. (2003). Aspects of the Syntax, the Pragmatics, and the Production of Code-Switching. Peter Lang.

Chan, B. H. S. (2008). Code-switching, word order and the lexical/functional category distinction. Lingua, 118(6), 777-809.

Cook, V. J. (1992). Evidence for multicompetence. Language learning, 42(4), 557-591.

Dashti, A. (2015). The role and status of the English language in Kuwait. English Today, 31(03), 28-33.

Dey, A., & Fung, P. (2014). A Hindi-English Code-Switching Corpus. In The 9th International Conference on Language Resources and Evaluation (LREC) (pp. 2410-2413).

Chiang, D., Diab, M. T., Habash, N., Rambow, O., & Shareef, S. (2006). Parsing Arabic Dialects. In EACL.

Elfardy, H., & Diab, M. T. (2012). Simplified guidelines for the creation of Large Scale Dialectal Arabic Annotations. In LREC (pp. 371-378).

Fleiss, J. L. (1971). Measuring nominal scale agreement among many raters. Psychological bulletin, 76(5), 378.

Gelderen, E., & MacSwan, J. (2008). Interface conditions and code-switching: Pronouns, lexical DPs, and checking theory. Lingua, 118(6), 765-776.

Grosjean, F. (1989). Neurolinguists, beware! The bilingual is not two monolinguals in one person. Brain and language, 36(1), 3-15.

Lederberg, A. R., & Morales, C. (1985). Code switching by bilinguals: Evidence against a third grammar. Journal of Psycholinguistic Research, 14(2), 113-136.

Li, Y., Yu, Y., & Fung, P. (2012). A Mandarin-English Code-Switching Corpus. In LREC (pp. 2515-2519).

Lyu, D. C., Lyu, R. Y., Chiang, Y. C., & Hsu, C. N. (2006, May). Speech recognition on code-switching among the Chinese dialects. In Acoustics, Speech and Signal Processing, 2006. ICASSP 2006 Proceedings.

2006 IEEE International Conference on (Vol. 1, pp. I-I). IEEE.

Maamouri, M., Bies, A., Kulick, S., Tabessi, D., & Krouna, S. (2012). Egyptian Arabic Treebank Pilot.

Maharjan, S., Blair, E., Bethard, S., & Solorio, T. (2015, June). Developing Language-tagged Corpora for Code-switching Tweets. In The 9th Linguistic Annotation Workshop held in conjuncion with NAACL 2015 (p. 72).

MacSwan, J. (1999). *A minimalist approach to intrasentential codeswitching*. New York and London: Garland, 1999. Language in Society, 30(02), 285-289.

MacSwan, J. (2000). The architecture of the bilingual language faculty: Evidence from intrasentential code switching. Bilingualism: language and cognition, 3(01), 37-54.

MacSwan, J. (2005). Codeswitching and generative grammar: A critique of the MLF model and some remarks on "modified minimalism". Bilingualism: language and cognition, 8(01), 1-22.

MacSwan, J. (2005). Remarks on Jake, Myers-Scotton and Gross's response: There is no "matrix language". Bilingualism: Language and cognition, 8(03), 277-284.

Myers-Scotton, C. (1993). Common and uncommon ground: Social and structural factors in codeswitching. Language in society, 22(04), 475-503.

Muysken, P. (2000). Bilingual speech: A typology of code-mixing (Vol. 11). Cambridge University Press.

Redouane, R. (2005). Linguistic constraints on codeswitching and codemixing of bilingual Moroccan Arabic-French speakers in Canada. In ISB4: Proceedings of the 4th International Symposium on Bilingualism (pp. 1921-1933).

Solorio, T., Blair, E., Maharjan, S., Bethard, S., Diab, M., Ghoneim, M., ... & Fung, P. (2014, October). Overview for the first shared task on language identification in code-switched data. In Proceedings of The First Workshop on Computational Approaches to Code Switching (pp. 62-72).

Solorio, T., & Liu, Y. (2008, October). Part-of-speech tagging for English-Spanish code-switched text. In Proceedings of the Conference on Empirical Methods in Natural Language Processing (pp. 1051-1060). Association for Computational Linguistics.

Zaidan, O. F., & Callison-Burch, C. (2011, June). The Arabic online commentary dataset: an annotated dataset of informal Arabic with high dialectal content. In Proceedings of the 49th Annual Meeting of the Association for Computational Linguistics: Human Language Technologies: short papers-Volume 2 (pp. 37-41). Association for Computational Linguistics.

## 11. Language Resource References

Linguistic Data Consortium. (2012). Egyptian Arabic Treebank DF Part 1 V2.0. Catalog No.: LDC2012E93.

Linguistic Data Consortium. (2012). Egyptian Arabic Treebank DF Part 2 V2.0. Catalog No.: LDC2012E98

Linguistic Data Consortium. (2012). Egyptian Arabic Treebank DF Part 3 V2.0. Catalog No.: LDC2012E89

Linguistic Data Consortium. (2012). Egyptian Arabic Treebank DF Part 4 V2.0. Catalog No.: LDC2012E99

Linguistic Data Consortium. (2012). Egyptian Arabic Treebank DF Part 5 V2.0. Catalog No.: LDC2012E107

Linguistic Data Consortium. (2012). Egyptian Arabic Treebank DF Part 6 V2.0. Catalog No.: LDC2012E125

Linguistic Data Consortium. (2012). Egyptian Arabic Treebank DF Part 7 V2.0. Catalog No.: LDC2013E12

Linguistic Data Consortium. (2013). Egyptian Arabic Treebank DF Part 8 V2.0. Catalog No.: LDC2013E21